\documentclass[]{interact}

\usepackage{subfigure}
\usepackage{graphicx}
\usepackage{algorithmic}
\usepackage{textcomp}
\usepackage{verbatim}
\usepackage{xcolor}
\usepackage{algorithm}  
\usepackage{booktabs}
\usepackage{amsmath}

\usepackage{epstopdf}
\usepackage[caption=false]{subfig}

\usepackage{natbib}
\bibpunct[, ]{(}{)}{;}{a}{}{,}
\renewcommand\bibfont{\fontsize{10}{12}\selectfont}

\theoremstyle{plain}

\theoremstyle{definition}

\theoremstyle{remark}

\begin{document}

\articletype{ARTICLE TEMPLATE}

\title{Framework Construction of an Adversarial Federated Transfer Learning Classifier}

\author{
\name{Hang Yi\textsuperscript{a}, Tongxuan Bie\textsuperscript{a} and Tongjiang Yan\textsuperscript{a}}
\affil{\textsuperscript{a}College of Science, China University of Petroleum (East China), Tsingtao, China
}}

\maketitle

\begin{abstract}
As the Internet grows in popularity, more and more classification jobs, such as IoT, finance industry and healthcare field, rely on mobile edge computing to advance machine learning. In the medical industry, however, good diagnostic accuracy necessitates the combination of large amounts of labeled data to train the model, which is difficult and expensive to collect and risks jeopardizing patients' privacy. In this paper, we offer a novel medical diagnostic framework that employs a federated learning platform to ensure patient data privacy by transferring classification algorithms acquired in a labeled domain to a domain with sparse or missing labeled data. Rather than using a generative adversarial network, our framework uses a discriminative model to build multiple classification loss functions with the goal of improving diagnostic accuracy. It also avoids the difficulty of collecting large amounts of labeled data or the high cost of generating large amount of sample data. Experiments on real-world image datasets demonstrates that the suggested adversarial federated transfer learning method is promising for real-world medical diagnosis applications that use image classification.
\end{abstract}


\section{Introduction}

Machine learning technology has been widely applied in real life in recent years, for example, in the Internet of Medical Things (IoMT). Health-related data, on the other hand, are existing in the form of several small datasets that are distributed around the globe due to major data privacy concerns in IoMT scenarios.

In this context, federated learning (FL) is gaining popularity because of its capacity to provide collaborative training while maintaining data privacy, as well as a solution to the problem of isolated data islands \citep{33}. The end-to-end diagnostic framework with automatic feature extraction may be simply established when the training and testing data are from the same distribution \citep{12} \citep{31}. However, the distributions of datasets in the real-world medical diagnosis industry vary by domain. One of the most typical issues is that labels are present in some datasets but are sparse or absent in others. This disparity in data categories is likely to impair the model's generalization capacity, resulting in differences between projected and actual results. The federated transfer learning (FTL) technique is presented as a solution to the problem \citep{14}. When the datasets contribute differently, federated transfer learning can train a good model in the domain with labeled data and then apply it to the domain with sparse data labels based on the relationship between the domains.

To minimize the impacts of domain shift in general transfer learning, numerous techniques have applied the Maximum Mean Discrepancy (MMD) loss \citep{9}. The deep Correlation Alignment (CORAL) approach advocated matching the mean and covariance of the two distributions with the same consideration \citep{24}.

Adversial-based domain adaptation approaches are what these two examples are referred to as. Peng et al. proposed to use the adversarial domain adaptation technique to address the domain shift effects in federated learning \citep{18}. The core notion of adversarial training is to continuously generate and learn adversarial samples in the process of network training \citep{8}. The Generative Adversarial Network (GAN) method is a generative deep model that pits two networks against one another: a generative model $G$ that captures the data distribution and a discriminative model $D$ that distinguishes between samples drawn from $G$ and images drawn from the training data by predicting a binary label. The networks are trained jointly using backpropagation on the label prediction loss in a mini-max fashion: simultaneously update $G$ to minimize the loss while also updating $D$ to maximize the loss. Therefore, the robustness of the training network will be improved.

However, the generative model's interpretability is poor, and its distribution cannot be articulated clearly. Gradient disappearance or non-convergence occurs during training, and there is no practical, direct, or noticeable technique for evaluating the effect of the generated model. Moreover, such generative models often face the problem of high cost of generating data in the practical application of FTL.

\begin{figure}
\centerline{\includegraphics[width=\columnwidth]{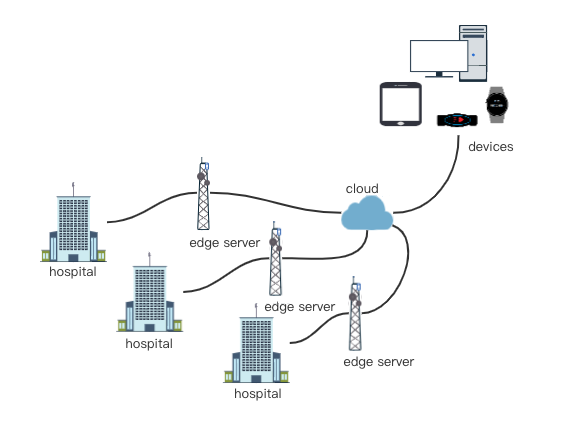}}
\caption{System architecture in healthcare diagnosis.}
\end{figure}

Since the FTL framework with adversarial ideas is ideally well suited for solving the problem of domain shifts due to sparsely labeled data in real smart medical scenarios, Therefore, we present an adversarial federated transfer learning (AFTL) classifier architecture for IoMT that employs current samples directly for training and allows the model trained on labeled datasets to migrate to sparsely labeled datasets for prediction. Unlike previous methods, this adversarial model we propose does not require the generation of new samples, but only the adversarial operation on the original data, which allows the interpretation of the model results and the effective evaluation of the effects based on the resolution of the domain bias phenomenon, while avoiding the higher costs required to produce the data.

Considering that the base of the world population is very large, the growth rate of doctors is far from that of patients. Therefore, such an AFTL framework is set up to allow patients to use their own wearable devices to learn the ability to diagnose medical images over the cloud server, so that they can use these devices to monitor their health status anytime and anywhere, breaking the limitation of distance and resources, while ensuring that their private data are not leaked.  In this system, labeled datasets from the source domain, such as medical data from hospital patients, are used to jointly train a nearly perfect model under the central server. Target domain, for example, the healthcare data stored in wearable devices or mobile devices, will obtain the trained model for classification prediction. In this method, instead of coming to the hospital, people can use personal devices to get a high-accuracy medical diagnosis classifier that ignores time and distance constraints and saves both human and material resources. It safeguards the privacy of medical data while also avoiding the significant costs associated with model generation. Figure 1. depicts the entire AFTL framework.

\section{Related works}

Machine learning has been successfully applied in medical diagnosis \citep{36}. For instance, the wearable devices can constantly monitor people's health problems at all times. It is mainly due to the extensive use of deep learning in image processing \citep{37}. In the field of medical diagnosis, the commonly used deep learning method is image recognition technology, which allows AI algorithms to learn more patient images than a human doctor can see in a lifetime to train a deep neural network \citep{38} that can determine whether a patient site is diseased or not \citep{39}. Wearable devices are developed to allow people who do not have regular health checkups to monitor their health status using medical image diagnostic models trained by deep learning \citep{40}.

In the medical field, many fundamental tasks can be done with deep learning technology as a way to reduce the workload of doctors, improve the efficiency of consultations, and increase the accuracy of medical diagnosis. Waston, an artificial intelligence system developed by the DeepQA initiative team led by IBM Principal Investigator David Ferrucci, has been used in clinical settings to provide doctors with advice on cancer diagnosis and treatment after studying a large number of textbooks and medical journals in the field of oncology over a four-year period \citep{41}. Watson will also simplify and standardize patient records, provide assistance in the collection and integration of laboratory and research data, and integrate the collected data into the Anderson Cancer Center's patient database so that these data can be analyzed in depth by advanced analytics \citep{42}. A team of researchers led by visiting professor Enda Wu from Stanford University demonstrated that a deep learning model can identify cardiac arrhythmias from electrocardiograms (ECGs), a method that can make a more reliable diagnosis of potentially lethal arrhythmias than cardiologists \citep{43}. For some areas with lower levels of medical care, this automated approach could improve cardiac diagnosis.

Although deep learning technology can help solve the problems of low efficiency and inaccurate results in medical diagnosis, it is difficult to collect and integrate data for training, as a large amount of treatment data of different cases of similar diseases and different conditions of the same patient are often stored in different medical institutions, and the willingness of each institution to share its own medical data is always limited. Also, personal medical data is very sensitive and involves important privacy information of users, many medical institutions do not have reasonable ways to apply and control complex medical data in the face of personal privacy protection and data security requirements.

Based on such considerations, federated learning is gradually gaining attention. The approach utilizes encryption algorithms to bypass the information barriers among medical institutions, and does not extract the original data of each participant, but transmits encrypted information through cryptographic protocols. This enables each medical institution to use the shared medical data for model training without exposing their original data \citep{44}. Tencent's Tianyin Lab and Microlife Bank jointly developed the "Stroke Onset Risk Prediction Model" based on the medical federal learning framework, which successfully cracked the problem of privacy protection in the medical industry and achieved accurate disease prediction with an accuracy rate of 80\% while protecting the privacy of data from different hospitals. In addition, through federal learning technology, the data resources of large hospitals can help small hospitals to improve the predictional accuracy of the model by 10-20\% \citep{45}.

However, in the post-promotion medical application scenario, the distribution of patients' health data collected by wearable devices is likely to be inconsistent with those stored in hospitals because of the different ways of storing data. The federal transfer learning technique gives a way to use data in a cross-medical institution scenario by using a model previously pre-trained in deep learning as the starting point for a new model to be applied to another new task \citep{46}. Specifically, a model firstly trained of fundus lesions using ImageNet, which can be used as a source domain, and then the model trained in that source domain is migrated to pneumonia diagnosis, enabling some of the parameters of the fundus lesion model to be shared by the pneumonia diagnosis model \citep{47}. W. Zhang proposes a prior distribution to indirectly bridge the domain gap when data from different clients cannot communicate and extracts client-invariant features while preserving data privacy \citep{48}. Generative Adversarial Networks (GAN), which have received much attention in recent years, can learn the distribution of data well and solve the shortage of annotated medical images \citep{8}. w. Zhang and X. Li propose a framework for fault diagnosis, which uses GAN to generate a large number of labeled training samples to solve the domain shift problem encountered in federated learning \citep{32}.

However, the large labeled datasets needed to train generative adversarial models are not always available in clinical practice because medical images require experts to acquire and label, and the number of patients with specific medical conditions may not be sufficient to create large datasets. This paper proposes a new adversarial learning method that does not require generating large training datasets to solve the problem of domain shift and sparse labels that exist in federated transfer learning, greatly reducing the cost of training and improving the accuracy of the FTL model in medical diagnosis, providing an implementable solution for wearable devices to monitor people's healthcare.

\section{Proposed method}

Let $\mathcal{D}^{s, i}=\left\{\left(\mathbf{x}_{j}^{s, i}, y_{j}^{s, i}\right)\right\}_{j=1}^{n^{s, i}}, i=1,2, \ldots, N$, denote the sample at the $i$-th source client, where $\mathbf{x}_{j}^{s, i} \in R$ and $y_{j}^{s, i}$ is the label of the $j$-th sample. $N$ is the number of source clients, ${n}^{s, i}$ denotes the number of samples in ${D}^{s, i}$. Also, let $\mathcal{D}^{t}=\left\{\left(\mathbf{x}_{j}^{t}\right)\right\}_{j=1}^{n^{t}}$ denote the sample at the target client, where $n^{t}$ is the number of the samples. The goal is to learn a classifier $C$ that can correctly classify target data into specified categories by utilizing the high-level features extracted by the feature extractor $F$, and a discriminator $D$ that cannot exactly distinguish whether the input data comes from the source or the target domain. 

For more intuitive expression, let $\theta_{F}$, $\theta_{C}$ and $\theta_{D}$ denote the parameters of $F$, $C$ and $D$, respectively. The optimization problem can be formulated as

\begin{equation}
\begin{aligned}
&\hat{\theta}_{F}=\arg \left\{\min _{\theta_{F}} L_{c}\left(\hat{\theta}_{C}, \theta_{F}\right), \max _{\theta_{F}} L_{d}\left(\hat{\theta}_{D}, \theta_{F}\right)\right\}, \\
&\hat{\theta}_{C}=\arg \min _{\theta_{C}} L_{c}\left(\theta_{C}, \hat{\theta}_{F}\right), \\
&\hat{\theta}_{D}=\arg \min _{\theta_{D}} L_{d}\left(\hat{\theta}_{F}, \theta_{D}\right),
\end{aligned}
\end{equation}

where $L_{c}$ denotes medical diagnosis error, and $L_{d}$ represents the domain prediction error. $\hat{\theta}_{F}$ , $\hat{\theta}_{C}$ and $\hat{\theta}_{D}$ denote the optimal
values of $\theta_{F}$, $\theta_{C}$ and $\theta_{D}$, respectively.

\subsection{Initialization process}

\begin{algorithm}[t]
\caption{Initialization}
\hspace*{0.02in} {\bf Input:} 
feature extractor, $N_{\text {epoch }}^{\text {init }}$, $\eta$ \\
\hspace*{0.02in} {\bf Output:} 
initialized parameters $\left\{\theta_{i}\right\}_{i=1}^{N}$ and $\theta_{t}$ 

\begin{algorithmic}[1]
\STATE select a representative client $l$ 
\FOR{each epoch $i=1,2, \ldots, N_{\text {epoch }}^{\text {init }}$}
 \STATE $\theta_{l} \leftarrow \theta_{l}-\eta \frac{\partial L_{i n i t}^{s, l}}{\partial \theta_{l}}$
\ENDFOR
\STATE initialize the models at the other clients with $\theta_{l}$ 
\end{algorithmic}
\end{algorithm}

The initialization process is shown in Figure 2. The representative $i$-th client is used to perform supervised learning on the labeled local data, and its classification loss based on cross-entropy can be expressed as
\begin{equation}
L_{c}^{s, i}=-\frac{1}{n^{s, i}} \sum_{j=1}^{n^{s, i}} \sum_{k=1}^{K} 1\left\{y_{j}^{s, i}=k\right\} \log \frac{e^{x_{j, k}^{i,c}}}{\sum\limits_{m=1}^{K} e^{x_{j, m}^{i,c}}},\label{eq}
\end{equation}
where $K$ denotes the number of possible diagnosis results, $x_{j, k}^{i,c}$ represents the $k$-th output value at the classifier $C$ of the $i$-th client which takes the $j$-th labeled sample as input.

Therefore, the optimized function of local initialization is defined as
\begin{equation}
\min L_{i n i t}^{s, i}=L_{c}^{s, i}. \label{eq}
\end{equation}

As the initialization of $i$-th client is finished, the initialized parameters will be broadcast to all other client models.

\subsection{Federated communication}

\begin{figure}
\centerline{\includegraphics[width=\columnwidth]{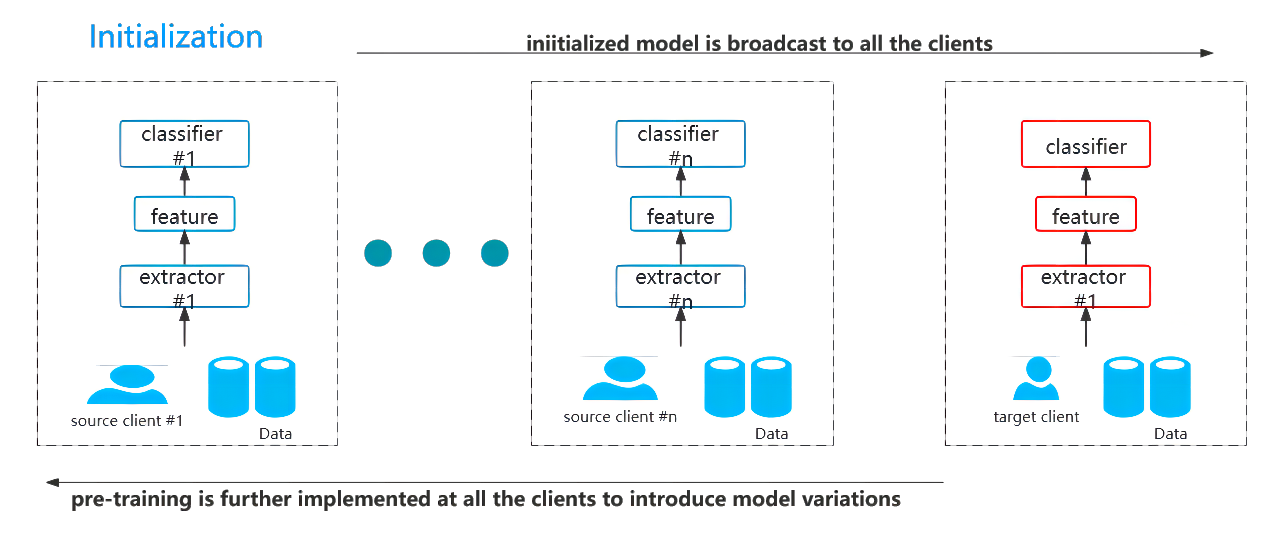}}
\caption{The initialization process.}
\end{figure}

\begin{figure}
\centerline{\includegraphics[width=\columnwidth]{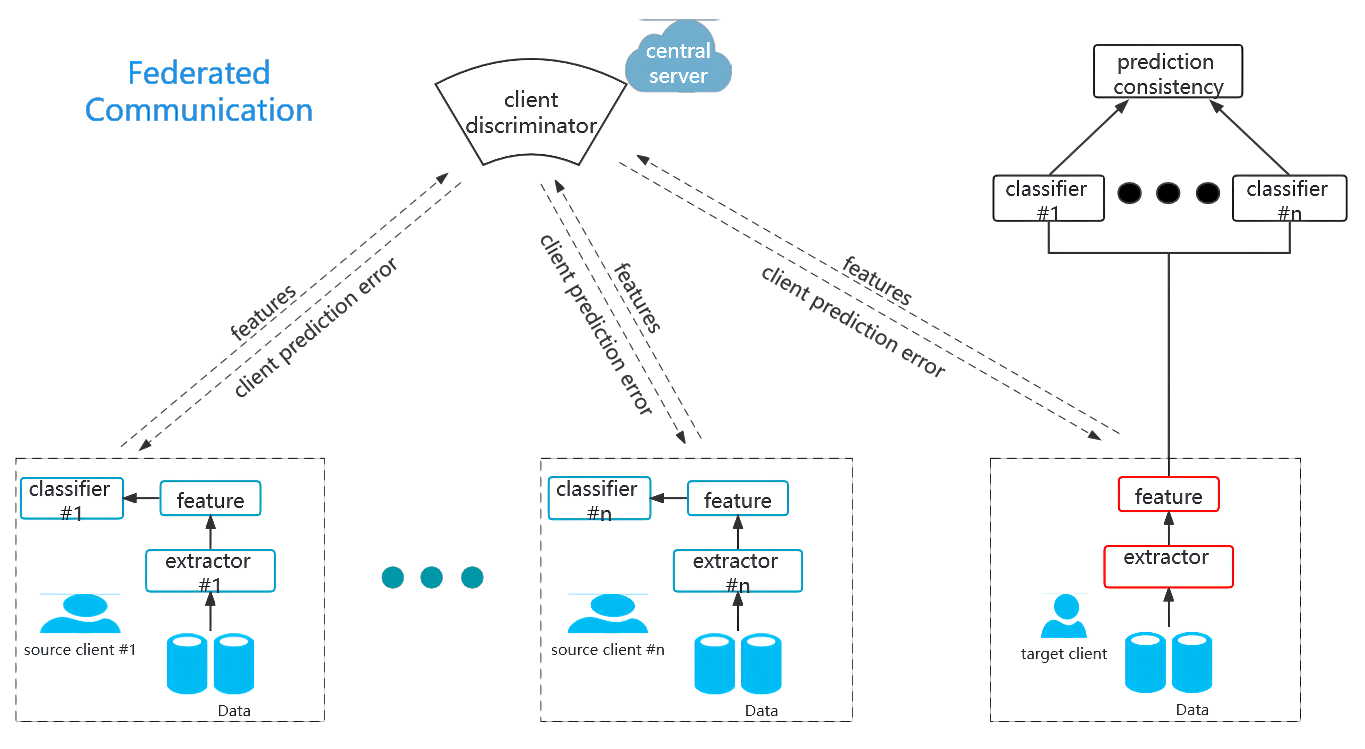}}
\caption{The phase of federated transfer learning.}
\end{figure}

At the federated communication process shown in Figure 3, the central server receives features uploaded from different clients, and the client (domain) discrimination error $L_{d}$ based on cross-entropy is formulated as
\begin{equation}
\begin{aligned}
L_{d} &=\sum_{i=1}^{K+1} L_{d}^{s, i}+L_{d}^{t}, \\
L_{d}^{s, i} &=-\frac{1}{n^{s, i}} \sum_{j=1}^{n^{s, i}} \sum_{k=1}^{K+1} 1\left\{d_{j}^{s, i}=k\right\} \log \frac{e^{x_{j, k}^{i, d}}}{\sum\limits_{m=1}^{K+1} e^{x_{j, m}^{i, d}}}, \\
L_{d}^{t} &=-\frac{1}{n^{t}} \sum_{j=1}^{n^{t}} \sum_{k=1}^{K+1} 1\left\{d_{j}^{t}=k\right\} \log \frac{e^{x_{j, k}^{t, d}}}{\sum\limits_{m=1}^{K+1} e^{x_{j, m}^{t, d}}},\label{eq}
\end{aligned}
\end{equation}
where $L_{d}^{s, i}$ and $L_{d}^{t}$ denote the client (domain) prediction loss at the $i$-th source and the target clients, respectively, $d_{j}^{s, i}$ denotes the client label of the $j$-th sample at the $i$-th source client, and $d_{j}^{t}$ is that of the $j$-th sample at the target client, $x_{j, k}^{i, \text { d }}$ and $x_{j, k}^{t, \text { d }}$ denote the $k$-th output at the client discriminator that take the $j$-th sample data at the $i$-th source and target clients respectively.

\begin{algorithm}[t]
\caption{Federated Communication}
\hspace*{0.02in} {\bf Input:} 
initialized $\left\{\theta_{i}\right\}_{i=1}^{N}$, $\theta_{t}$, $N_{\text {round }}$, $N_{\text {epoch }}^{\text {fed }}$ \\
\hspace*{0.02in} {\bf Output:} 
trained parameters $\left\{\theta_{i}\right\}_{i=1}^{N}$, $\theta_{t}$ 

\begin{algorithmic}[1]
\STATE Local client:
\FOR{each training round $j=1,2, \ldots, N_{\text {round }}$}
 \FOR{each training round $j=1,2, \ldots, N$}
  \STATE receive source client  prediction loss from central server
  \STATE update model with $\theta_{F}^{s, i} \leftarrow \theta_{F}^{s, i}-\eta\left(\frac{\partial L_{c}^{s, i}}{\partial \theta_{F}^{s, i}}-\frac{\partial L_{d}^{s, i}}{\partial \theta_{F}^{s, i}}\right)$ and $\theta_{C}^{s, i} \leftarrow \theta_{C}^{s, i}-\eta\left(\frac{\partial L_{c}^{s, i}}{\partial \theta_{C}^{s, i}}+\frac{\partial L_{p}}{\partial \theta_{C}^{s, i}}\right)$
  \STATE upload features of source client data to central server 
 \ENDFOR
 \STATE receive target client prediction loss from central server
 \STATE update target client model with $\theta_{F}^{t} \leftarrow \theta_{F}^{t}-\eta\left(-\frac{\partial L_{d}^{t}}{\partial \theta_{F}^{t}}\right)$
 \STATE Upload features of target client data to central server
\ENDFOR
\STATE Central server: 
\FOR{$j=1,2, \ldots, N_{\text {round }}$}
 \STATE receive data features from all clients
 \STATE update client discriminator with $\theta_{D} \leftarrow \theta_{D}-\eta \frac{\partial L_{d}}{\partial \theta_{D}}$
 \STATE send discrimination loss to all clients
\ENDFOR
\end{algorithmic}
\end{algorithm}

Note that the source classifiers are different. To enhance the generalization ability of the model and reduce the risk of overfitting, the diagnostic loss of different source classifiers $L_{p}$ should be minimized as 
\begin{equation}
\begin{aligned}
\min L_{p} &=\frac{1}{n^{t} N} \sum_{j=1}^{n^{t}} \sum_{i=1}^{N}\left\|C^{s,i}\left(F^{t}\left(\mathbf{x}_{j}^{t}\right)\right)-\bar{C}\left(F^{t}\left(\mathbf{x}_{j}^{t}\right)\right)\right\|, 
\end{aligned}\label{eq}
\end{equation}
where $C^{s, i}$ is the classifier at the $i$-th source client. The mean vector of the predictions $\bar{C}\left(F^{t}\left(\mathbf{x}_{j}^{t}\right)\right)$ is defined as
\begin{equation}
\bar{C}\left(F^{t}\left(\mathrm{x}_{j}^{t}\right)\right)=\frac{1}{N} \sum_{i=1}^{N} C^{s, i}\left(F^{t}\left(\mathbf{x}_{j}^{t}\right)\right). \label{eq}
\end{equation}

Through the steps of integrated supervised learning, feature mapping and adversarial learning, etc., the optimization problem in AFTL can be expressed as
\begin{equation}
\begin{aligned}
{\theta}_{F}^{s, i} &=\arg \left\{\min _{\theta_{F}^{s, i}} L_{c}^{s, i}, \max _{\theta_{F}^{s, i}} L_{d}^{s, i}\right\}, \\
{\theta}_{F}^{t} &=\arg  \max _{\theta_{F}^{t}} L_{d}^{t}, \\
{\theta}_{C}^{s, i} &=\arg \min  _{\theta_{C}^{s, i}} L_{c}^{s, i}+L_{p}, \\
{\theta}_{D} &=\arg \max _{\theta_{D}} L_{d},
\end{aligned}\label{eq}
\end{equation}
where $\theta_{F}^{s, i}$ and $\theta_{C}^{s, i}$ denote the parameters of the feature extractor and classifier at the $i$-th source client respectively. $\theta_{F}^{t}$ is the parameter of the feature extractor at the target client. 

The input of the image classifier and client discriminator are all from the feature extractor, however, the discriminator aims to maximize the domain classification loss that confuses source and target clients while the classifier is leading to minimize the image classification loss so that classification accuracy of images can be improved:
\begin{equation}
\theta_{F}^{s, i}=\arg \left\{\min _{\theta_{F}^{s, i}} L_{c}^{s, i}, \max _{\theta_{F}^{s, i}} L_{d}^{s, i}\right\}. \label{eq}
\end{equation}

This leads to the fact that the direction of the discrimination loss gradient is opposite to that of the classification loss gradient when parameters update at the feature extractor. To avoid staged optimization, a gradient reversal layer is embedded between the feature extractor and the discriminator. The gradient of discrimination loss will be automatically opposite before backpropagating to the feature extractor. 

With this idea of adversarial learning, the parameters at above process can be updated as
\begin{equation}
\theta_{F}^{s, i} \leftarrow \theta_{F}^{s, i}-\eta\left(\frac{\partial L_{c}^{s, i}}{\partial \theta_{F}^{s, i}}-\frac{\partial L_{d}^{s, i}}{\partial \theta_{F}^{s, i}}\right). \label{eq}
\end{equation} 

In summary, the update of parameters in AFTL can be expressed as
\begin{equation}
\begin{aligned}
&\theta_{F}^{s, i} \leftarrow \theta_{F}^{s, i}-\eta\left(\frac{\partial L_{c}^{s, i}}{\partial \theta_{F}^{s, i}}-\frac{\partial L_{d}^{s, i}}{\partial \theta_{F}^{s, i}}\right), \\
&\theta_{F}^{t} \leftarrow \theta_{F}^{t}-\eta(-\frac{\partial L_{d}^{t}}{\partial \theta_{F}^{t}}), \\
&\theta_{C}^{s, i} \leftarrow \theta_{C}^{s, i}-\eta\left(\frac{\partial L_{c}^{s, i}}{\partial \theta_{C}^{s, i}}+\frac{\partial L_{p}}{\partial \theta_{C}^{s, i}}\right), \\
&\theta_{D} \leftarrow \theta_{D}-\eta (-\frac{\partial L_{d}}{\partial \theta_{D}}), \label{eq}
\end{aligned}
\end{equation}
where $\eta$ is the learning rate. The detailed training implementations of the proposed method are presented in Algorithms 1 and 2.

\subsection{Testing process}
After training, each source client has trained its unique image classifier. To improve the consistency of prediction under the different classifiers, the test image of the target client is used for testing, and then select the classification result with the most votes as the prediction of the image at the target client.

\section{Experiment}

We evaluate the AFTL method for unsupervised classification tasks across 2 different data settings. 10 source clients and 1 target client are utilized. On the MNIST dataset whose data samples have a similar distribution, we test the suggested method for a relatively straightforward knowledge transfer, whereas the Office dataset has data samples from three different distributions aimed at sophisticated learning. Example images from all experimental datasets are provided in Figure 4. For comparison, we employed transfer learning (TL), adversarial transfer learning (ATL) and federated transfer learning (FTL) which are all sophisticated classification approaches. The performance of the AFTL method is also verified in the following experiment.

\begin{figure}
\centerline{\includegraphics[width=\columnwidth]{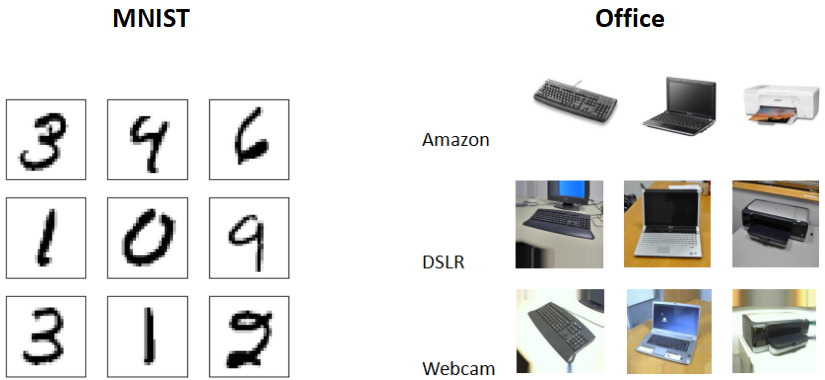}}
\caption{Example images from two datasets mentioned above.}
\end{figure}

\subsection{MNIST Dataset}
The MNIST dataset contains 60,000 training images and 10,000 testing images that were anti-aliased and normalized to fit into a 28x28 pixel bounding box. We give 10 source clients 15000 samples for model training and 1000 samples to the target client for testing. The image features are extracted using the Conv2d algorithm. The experimental results are shown in Table 1. Easier MNIST datasets  made the performance better than the current state-of-the-art methods.

\subsection{Office Dataset}
The benchmark Office visual dataset contains 4,110 photos from 31 different classes in three different domains: amazon, webcam, and DSLR. We used 2817 amazon photos for source clients and 795 webcam images for target clients in the data settings. The goal is to compare AFTL to other algorithms based on every labeled example from Amazon and every unlabeled example from the webcam.

\begin{table*}
\tbl{Test accuracy on several transfer learning among MNIST and Office31} 
{\begin{tabular}{p{85pt} p{85pt} p{25pt} }
\hline
Method&MNIST&Office31\\
\hline
TL    & 65.7\% & 75.3\%   \\
ATL   & 75.2\% & 89.6\%         \\
FTL    & 85.8\% & 79.2\%         \\
AFTL   & 93.4\% &  90.1\%      \\
\hline
\end{tabular}}
\end{table*}

\subsection{Convergence Details}
In this study, the accuracy of the classification model and the value of the loss function are selected as the analysis indicators. The effect of the model is proportional to the accuracy rate, while inversely related to the value of the loss function.

\begin{figure}
    \begin{minipage}[t]{0.5\linewidth}
        \centering
        \includegraphics[width=\textwidth]{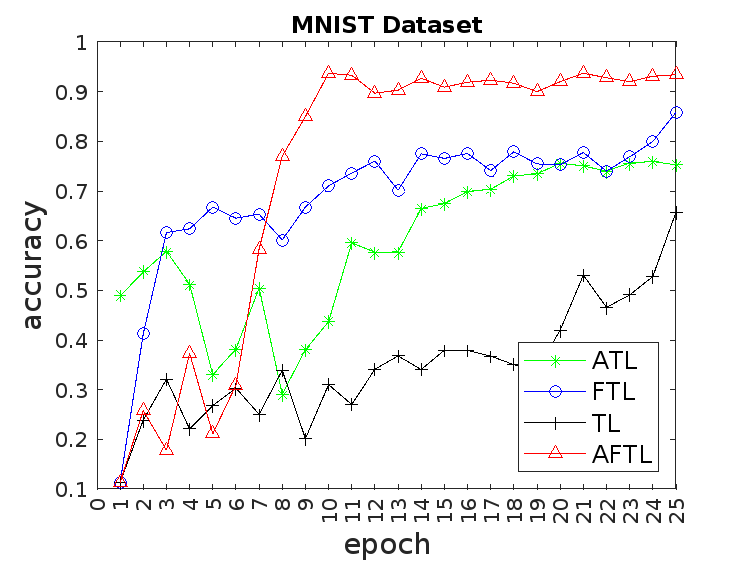}
        \centerline{(a)}
    \end{minipage}%
    \begin{minipage}[t]{0.5\linewidth}
        \centering
        \includegraphics[width=\textwidth]{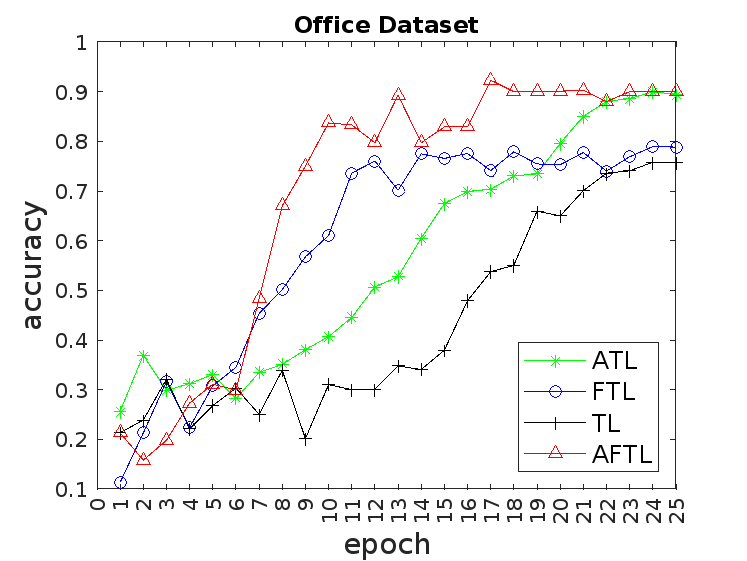}
        \centerline{(b)}
    \end{minipage}

    \begin{minipage}[t]{0.5\linewidth}
        \centering
        \includegraphics[width=\textwidth]{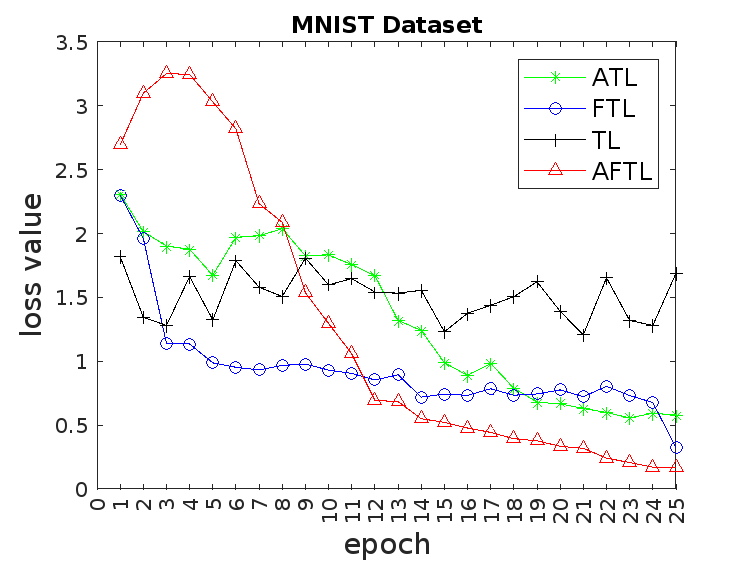}
        \centerline{(c)}
    \end{minipage}%
    \begin{minipage}[t]{0.5\linewidth}
        \centering
        \includegraphics[width=\textwidth]{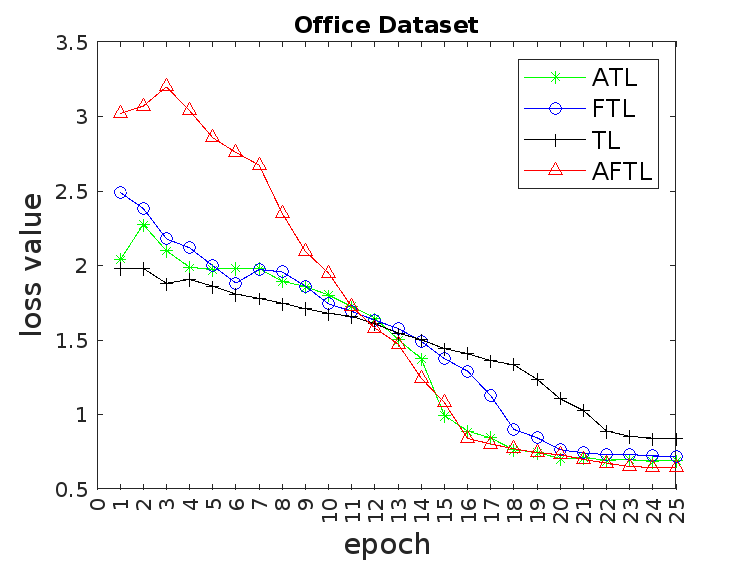}
        \centerline{(d)}
    \end{minipage}
    \caption{Test accuracy on the target client and average loss value on the source clients.}
\end{figure}

In our experiment, the batch size is selected as 100 and the number of iterations is set up at 100. Figure 5.(a)(b) indicate that in the early stage of training, the accuracy is very low because the model is still unfamiliar with the sample. The accuracy of the test grows as the number of iterations increases, reaching a fixed number at the 30-th iteration and stabilizing near this figure. If the model's expressive ability is too high, it will learn some non-common characteristics that can only satisfy the training sample, resulting in a reduction in test accuracy. Clearly, the findings of our tests did not lead to this circumstance.

The cross-entropy loss is used to evaluate the model's performance in this article's multi-classification challenge. Figure 5.(c)(d) show how the model training improves as the number of iterations grows, and the loss function on the classifier in the test phase rapidly reduces. Near the 30-th time, it also hits the minimal value. The classification model has reached a good point at this point. As a consequence, we can say that our approach is convergent since it allows the prediction accuracy to stabilize at a high level while reducing the classification result loss.

\begin{figure}
    \begin{minipage}[t]{0.5\linewidth}
        \centering
        \includegraphics[width=\textwidth]{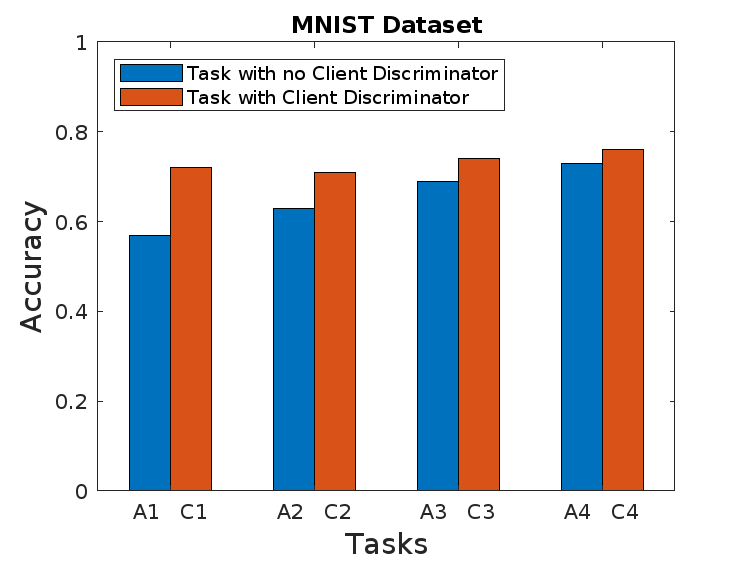}
        \centerline{(a)}
    \end{minipage}%
    \begin{minipage}[t]{0.5\linewidth}
        \centering
        \includegraphics[width=\textwidth]{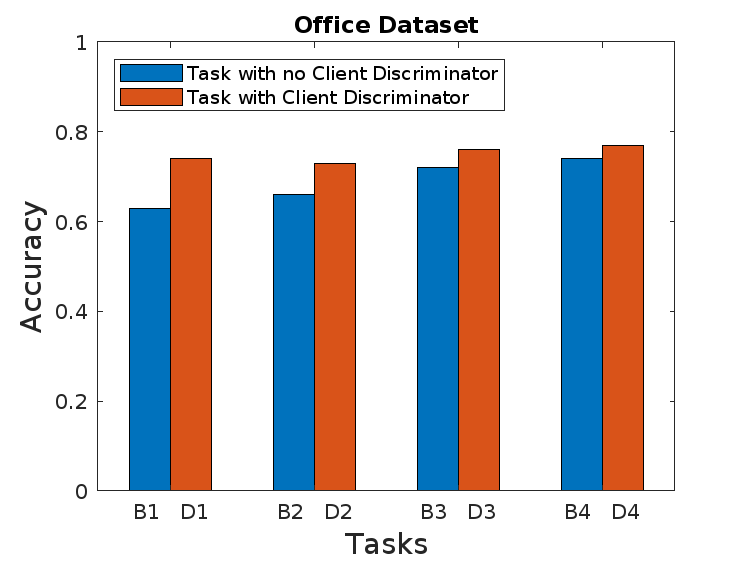}
        \centerline{(b)}
    \end{minipage}
    \caption{Performance on the AFTL algorithm.}
\end{figure}

We also set up a set of ablation experiments as a way to determine the impact of the new performance on the AFTL algorithm as shown in Table 2. and Table 3. When we remove the client discriminator on the central server, the accuracy of the test results is generally lower than that with the discriminator, indicating that the effect of our proposed client discriminator is significant as shown in Figure 6. Meanwhile, in the method without the discriminator, the accuracy of the test results increases with the increase of the sample size, while with the discriminator included, the accuracy of the test results is not significantly affected by the sample size, indicating that the proposed method is robust to the sample size, further demonstrating that the method can guarantee the training effect without the need to generate new sample data.

\begin{table*}
\tbl{The settings of learning on the MNIST dataset.} 
{\begin{tabular}{p{45pt} p{75pt} p{65pt} p{65pt} }
\hline
Task name&AFTL with a Client Discriminator&Client number&Sample number\\
\hline
A1  & no & 5 & 200  \\
A2  & no & 10 & 100   \\
A3  & no & 5 & 800        \\
A4  & no & 10 & 400    \\
C1  & yes & 5 & 200  \\
C2  & yes & 10 & 100        \\
C3  & yes & 5 & 800       \\
C4  & yes & 10 & 400    \\
\hline
\end{tabular}}
\end{table*}

\begin{table*}
\tbl{The settings of learning on the Office dataset.} 
{\begin{tabular}{p{45pt} p{75pt} p{65pt} p{65pt} }
\hline
Task name&AFTL with a Client Discriminator&Client number&Sample number\\
\hline
B1  & no & 5 & 200  \\
B2  & no & 10 & 100   \\
B3  & no & 5 & 800        \\
B4  & no & 10 & 400    \\
D1  & yes & 5 & 200  \\
D2  & yes & 10 & 100        \\
D3  & yes & 5 & 800       \\
D4  & yes & 10 & 400    \\
\hline
\end{tabular}}
\end{table*}

\section{Conclusion}
This article aimed to overcome the problems of data shift and privacy leakage in federated transfer learning by merging a robust neural network framework based on adversarial idea. Experiments on two separate data sets later verified that our methodology outperforms the competition.

In this paper, we propose an AFTL framework for the medical diagnosis that performs adversarial operations without generating new large amounts of data, which can solve the problems of data privacy and domain shift faced by medical diagnosis scenarios, while avoiding the high cost in previous generative adversarial methods and also improving the accuracy of training results. The method proves to be experimentally superior to other methods, while the performance of our proposed new component also proves to be significantly important. The method further advances the application of AI technology in the field of medical diagnosis.

\end{document}